\documentclass[12pt]{elsarticle}

\usepackage{setspace}
\usepackage{booktabs}
\usepackage{threeparttable}
\usepackage{longtable}
\usepackage[table,xcdraw]{xcolor}
\usepackage{multirow}
\usepackage{float}
\usepackage{subcaption}
\usepackage{parskip}
\usepackage{pdflscape}
\usepackage{graphicx}
\usepackage{tikz} 
\usepackage{lineno}
\usepackage{enumitem}
\usepackage{xcolor}
\usepackage{adjustbox}
\usepackage{tabularx}
\usepackage[hidelinks]{hyperref}
\usetikzlibrary{trees, positioning, arrows.meta}
\usetikzlibrary{arrows,decorations.pathmorphing,backgrounds,fit,positioning,shapes.symbols,chains}
\setcitestyle{square}

\begin{document}


\begin{frontmatter}



\title{Tell Me a Story! Narrative-Driven XAI with Large Language Models}



\author[inst1]{David Martens}
\author[inst1]{James Hinns}
\author[inst1]{Camille Dams}
\author[inst1]{Mark Vergouwen}
\author[inst2]{Theodoros Evgeniou}

\affiliation[inst1]{organization={University of Antwerp, Department of Engineering Management},
            addressline={Prinsstraat 13}, 
            city={Antwerp},
            postcode={2000}, 
            country={Belgium}}
\affiliation[inst2]{organization={INSEAD},
            addressline={Boulevard de Constance}, 
            city={77305 Fontainebleau},
            country={France}}
\begin{abstract}
In many AI applications today, the predominance of black-box machine learning models, due to their typically higher accuracy, amplifies the need for Explainable AI (XAI). Existing XAI approaches, such as the widely used SHAP values or counterfactual (CF) explanations, are arguably often too technical for users to understand and act upon. To enhance comprehension of explanations of AI decisions and the overall user experience, we introduce XAIstories, which leverage Large Language Models to provide narratives about how AI predictions are made: SHAPstories do so based on SHAP explanations, while CFstories do so for CF explanations. We study the impact of our approach on users' experience and understanding of AI predictions. Our results are striking: over 90\% of the surveyed general audience finds the narratives generated by SHAPstories convincing. Data scientists primarily see the value of SHAPstories in communicating explanations to a general audience, with 83\% of data scientists indicating they are likely to use SHAPstories for this purpose.  In an image classification setting, CFstories are considered more or equally convincing as the users' own crafted stories by more than 75\% of the participants. CFstories additionally bring a tenfold speed gain in creating a narrative. We also find that SHAPstories help users to more accurately summarize and understand AI decisions, in a credit scoring setting we test, correctly answering comprehension questions significantly more often than they do when only SHAP values are provided. The results thereby suggest that XAIstories may significantly help explaining and understanding AI predictions, ultimately supporting better decision-making in various applications.
\end{abstract}



\begin{keyword}
Explainable AI \sep Counterfactual Explanations \sep SHAP \sep Large Language Models \sep XAIstories 
\end{keyword}

\makeatletter
\def\ps@pprintTitle{%
  \let\@oddhead\@empty
  \let\@evenhead\@empty
  \let\@oddfoot\@empty
  \let\@evenfoot\@empty
}
\makeatother

\end{frontmatter}


\section{Introduction}
\label{Intro}
A key problem due to the growing adoption of black box Machine Learning models is to improve the transparency of such models \citep{Adadi2018, Molnar2022}. Explainable AI (XAI) aims at making the underlying decision mechanisms of black box models understandable to both system developers and end users of the model. Such explanations have been proven crucial for justifying model predictions towards data subjects, understanding misclassifications, improving a model’s reliability in real-life applications, and increasing users’ trust in the model \citep{Adadi2018, Gohel2021}. However, state-of-the-art XAI methods do not generate concise, coherent, natural language-based explanations, with a narrative that can possibly further help users. 

\begin{figure}[H]
\begin{center}
\hspace*{-2cm}
\includegraphics[width=1.3\textwidth]{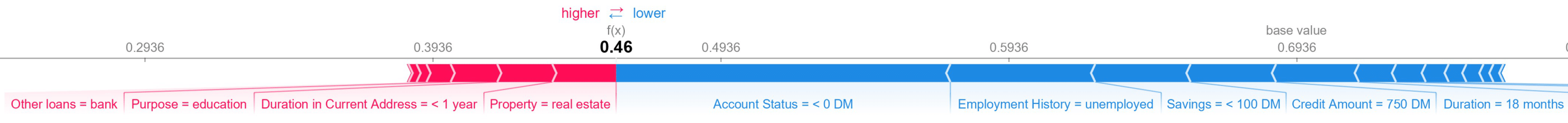} 
\caption{Example of a SHAP explanation for a rejected (0) loan application.}
\label{fig:introexample}
\end{center}
\end{figure}

\noindent For instance, consider the case of the popular SHAP (SHapley Additive exPlanations) method \citep{Lundberg2017}. The goal of SHAP is to explain the prediction score of an instance by computing the contribution of each feature to the prediction. 
An example of a SHAP explanation for a loan application that is rejected by an AI model is given in Figure~\ref{fig:introexample}. The features include various credit and loan applicant characteristics. 
The reader is invited to interpret the figure and create a narrative for the applicant about why the AI model predicts poor creditworthiness - we revisit this example in Section \ref{Results}. 

\noindent This small exercise shows the nuisance that many data scientists struggle with, when they need to convey a story to a lay user on why the AI made a prediction. Even though SHAP values can support both data scientists and end users in understanding predictions of black-box models, they simply do not generate natural language-based explanations. On top of that, SHAP values also simply lack conciseness since they attribute significance to all individual features, which often goes into the dozens or even hundreds \citep{Molnar2022}. 
 
\noindent In this paper, we address the need to generate natural language explanations that provide 
a narrative that interprets an XAI explanation\footnote{The curious reader may already look at Table~\ref{table:appendix_SHAPstories_question_student} or~\ref{table:appendix_SHAPstories_question_loan} for a generated XAIstory.}. Our contributions are threefold:
\begin{enumerate}
    \item We introduce and motivate a new XAI approach, thereby opening up a new research area into generating, evaluating and deploying such an approach.
    \item We provide a first open-source implementation (prompt-based with GPT-4) to generate narratives for two commonly used XAI methods: counterfactuals and SHAP values.
    \item We demonstrate, using four survey studies, that the proposed approach leads to an improved user experience (in terms of convincingness, ease of use, confidence and likeliness to use), as well as improved users' comprehension of the explanations provided. 
\end{enumerate}


\noindent The paper is structured as follows. Section \ref{literature} positions this research in the current literature, which also motivates the need for XAIstories. Section \ref{methodology} details the definition and methodology behind XAIstories. Next, Section \ref{Results} describes the output of \mbox{XAIstories} and the results derived from user study surveys, while Section \ref{conclusion} concludes the paper.

\section{Related literature} \label{literature}
An explanation is an attempt to convey the internal state or logic of an algorithm that leads to a decision~\cite{wachter2017}. Prior work has suggested that when users do not understand the workings of the classification model, they become more skeptical and reluctant to use the model, even if the model is known to improve decision performance~\cite{Kayande09}. This brings us to the main motivation for explanations: trust that the model has learnt the right patterns~\cite{martens2022data}. There are two sorts of explanation procedures for prediction models: global explanations, which provide an understanding for the model's overall behavior across all (training) instances~\cite{martens2022data,Martens2014}, and instance-level explanations, which explain the model's prediction for a single instance~\cite{Molnar2022,Martens2014,Ribeiro2016}.
As most often we need to explain single predictions, the focus of current XAI research is largely on instance-based explanations, with feature attribution and counterfactual (CF) methods being two popular solutions~\cite{Molnar2022,martens2022data}. 

A CF explanation provides a minimum set of evidence present in the data instance to be explained (also called the factual), such that removing that evidence would change the decision~\cite{Martens2014,wachter2017}. In recent years, various techniques have been proposed to create such explanations, for tabular~\cite{wachter2017}, image~\cite{Vermeire2022}, and textual data~\cite{martens2022data,Kim2020}. LIME and SHAP are feature attribution methods that provide a set of feature-value combinations with weights that indicate their importance in the prediction~\cite{Lundberg2017,martens2022data}, applicable to various data types.

\noindent The literature already points to the need for \textit{narrative-based} solutions. First, there is a mismatch between current technical XAI solutions and human decision making, also known as the \textit{`inmates running the asylum'} phenomenon~\citep{Miller2017}. This mismatch arises because AI researchers are often creating solutions that serve their own needs rather than those of the intended non-expert users~\cite{yang2023}. While developers and data scientists may seek detailed, algorithmic transparency to pinpoint errors or refine the model, end users and domain experts often require a clear, comprehensible \textit{narrative} that explains decisions they can understand. 

\noindent Second, to also overcome the issue above, narratives from XAI outputs are typically already created manually, either by a domain expert or by the users themselves, with less ease and control in the latter case. For example, De Lange et al (2022) provide narratives to describe SHAP waterfall plots in financial intermediation \citep{DeLange2022}, while Zhou et al. (2023) enhance SHAP explanations with narrative descriptions in a fraud detection application~\citep{Zhou2023}. Similarly, narratives are important in employment and real estate applications, as shown by De Oliveira et al. (2023) and Varun et al. (2020), helping to make AI models easier to adopt and more practical to use \citep{DeOliveira2023, Provost2020}. 

\noindent Third, the fields of communication theory and psychology also point to the benefits of a narrative-based approach to explanations. For example, narratives are naturally more accessible and memorable, making them particularly effective when communicating scientific evidence to a non-expert audience~\citep{Dahlstrom2014}. Research also suggests that narrative explanations are processed more quickly than logical-scientific descriptions by non-expert audiences, regardless of their prior knowledge of the topic \citep{Graesser2003}. Moreover, narratives tend to be more engaging and persuasive, enhancing trust and understanding of AI models among non-experts \citep{Dahlstrom2014, Bullock2021}.

\noindent There are also legal reasons for the use of narratives when explaining AI decisions. For instance, under European GDPR Article 12 and Recital 58, the principle of transparency requires ``\textit{concise, easily accessible and easy to understand, and that clear and plain language and, additionally, where appropriate, visualisation be used}''~\citep{Walke2023, Wulf2024}.  

Our work on XAIstories is also in line with the general research agenda in human-AI interactions~\cite{dssstorey2024}, which emphasizes the need for designs that \textit{``improve decision-making performance’’} by effectively aligning AI capabilities with human decision making processes. By automatically developing narrative explanations that make AI decisions more comprehensible, our research directly contributes in this direction.

Finally, we note there is recent related work developing at the intersection of XAI and LLMs. For example, TalkToModel is an interactive dialogue system that chooses an XAI method and returns an explanation in textual form~\citep{Slack2023}. The scope of TalkToModel, and other such work, is however different: it interprets a question, maps it to appropriate explanations, and generates a textual response to the question. The last step does not create narratives about AI predictions, but rather simpler responses such as how important a feature is, what the 
most important features are, or what a counterfactual explanation may be. XAIstories, on the other hand, provide more complete narratives about AI predictions.

\section{Methodology} \label{methodology}

\subsection{Narratives Generation}
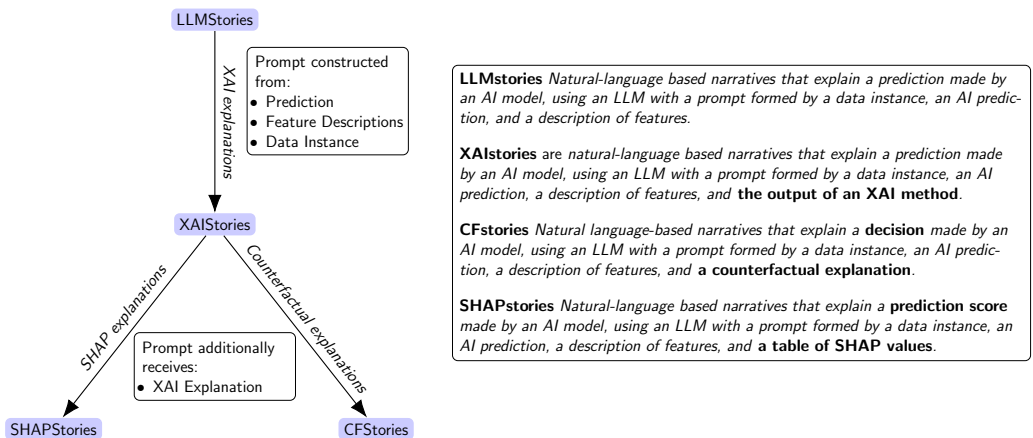
\begin{figure}
\centering
\begin{adjustbox}{max width=\textwidth, max height=\textheight}
    \begin{tikzpicture}[
  edge from parent/.style={draw, thick, ->, >={Latex[scale=2]}, font=\sffamily, midway, sloped},
  sibling distance=20em,
  level distance=12em,
  every node/.style={font=\sffamily, rounded corners, fill=blue!20, text centered, anchor=north},
  edge label/.style={font=\sffamily\itshape, fill=none, rounded corners=0pt, text=black}
]

\node (root) {LLMStories}
  child {node (xai) {XAIStories}
    child {node {SHAPStories}
      edge from parent node[edge label, above] {SHAP explanations}}
    child {node {CFStories}
      edge from parent node[edge label, above] {Counterfactual explanations}}
    edge from parent node[edge label, above] {XAI explanations}}
  ;

\node[anchor=west, text width=10em, fill=white, draw=black, rounded corners, align=left, below=0 of root, yshift=-1em, xshift=7em, inner sep=0pt] (bulletpoints-root) {
\\
\hspace{0.5em}Prompt constructed \\
\hspace{0.5em}from:
  \begin{itemize}[nosep, left=2pt, align=left]
    \item Prediction
    \item Feature Descriptions
    \item Data Instance \vspace{0.5em}  
  \end{itemize}
};

\node[anchor=south, text width=10em, fill=white, draw=black, rounded corners, align=left, below=0 of xai, yshift=-6em, xshift=0em, inner sep=0pt] (bulletpoints-xai) {
\\
\hspace{0.5em}Prompt additionally \\
\hspace{0.5em}receives: \\
\begin{itemize}[nosep, left=2pt, align=left]
    \item XAI Explanation \vspace{0.3em}
  \end{itemize}
};

\node[anchor=west, align=left, text width=35em, fill=white, draw=black, rounded corners, inner sep=5pt, right=12em of root, yshift=-12em] (definitions) {
    \textbf{LLMstories} \textit{Natural-language based narratives that explain a prediction made by an AI model, using an LLM with a prompt formed by a data instance, an AI prediction, and a description of features.} \\ [1em]
    \textbf{XAIstories} are \textit{natural-language based narratives that explain a prediction made by an AI model, using an LLM with a prompt formed by a data instance, an AI prediction, a description of features, and \textbf{the output of an XAI method}.} \\ [1em]
    \textbf{CFstories} \textit{Natural language-based narratives that explain a \textbf{decision} made by an AI model, using an LLM with a prompt formed by a data instance, an AI prediction, a description of features, and \textbf{a counterfactual explanation}.} \\ [1em]
    \textbf{SHAPstories} \textit{Natural-language based narratives that explain a \textbf{prediction score} made by an AI model, using an LLM with a prompt formed by a data instance, an AI prediction, a description of features, and \textbf{a table of SHAP values}.}
};


\end{tikzpicture}
\end{adjustbox}
\caption{Taxonomy diagram and definitions for LLM, XAI, SHAP and CF-stories.}\label{fig:taxonomy}
\end{figure}

We combine XAI methods, specifically SHAP values or CF explanations, with LLMs. In particular, given a machine learning model, an input, the output of the model for that input, and an XAI explanation for that model output, we construct different types of prompts that we provide to an LLM which then provides a narrative. We study how different prompt strategies affect the narratives, and the reactions of users to these narratives, using four surveys. 

\textbf{Prompt Types}: The methods we present involve building specialised prompts to an LLM so that it can explain an AI prediction. As shown in Figure \ref{fig:taxonomy} we present two types of prompts: for LLMstories and for XAIstories. Both are agnostic to both models to be explained and LLMs used to produce narratives.
\begin{itemize}
    \item \textit{LLMstories} include the input instance and prediction to be explained, along with a description of the features. 
    \item \textit{XAIstories} additionally include an XAI explanation (SHAP or CF) - see below for more details.
\end{itemize}
We then implement two special cases of XAIstories; SHAPstories and CFstories, which use SHAP and Counterfactuals as their respective XAI explanations to integrate into the prompt.

\textbf{Large Language Model (LLM)}: 
We use GPT-4 to generate XAIstories, as it is currently widely recognized as a well-performing LLM able to process both textual and visual inputs~\cite{yang2024,bubeck2023}. Furthermore, it requires no fine-tuning for specific tasks, is accessible via an API, and has integrated plug-ins. As we discuss in Section~\ref{methodologyXAImethods}, we use these plugins as part of the pipeline to generate CFstories.

We also experimented with Google's Gemini and Meta's Llama LLMs. Based on initial experiments, we found those narratives to be less convincing than the ones generated by GPT-4. This is only anecdotal evidence, and future  studies can assess to what extent advances in LLMs can further improve our approach and findings. From that perspective, our results are a lower bound for the performance of XAIstories in the future. To facilitate the use of different LLMs to generate and test XAIstories, we wrote generic wrappers where an API to any LLM can be inserted in order to prompt an LLM automatically. These wrappers were tested with GPT-4, Gemini (7b) and Llama 3 (70b instruct), but can be extended to other LLMs as desired. 
 All of these models were tested via APIs, however it is also possible to extend the wrappers with local models.
These implementations are all found on our GitHub: \url{https://github.com/ADMAntwerp/XAIstories}.

\subsection{Data and implementation}\label{methodologyXAImethods}

\textbf{Image Classification:} In the context of image classification, we study narratives for counterfactual explanations. We use \textbf{S}emantic \textbf{C}ounterfactuals for \textbf{A}ccurate \textbf{P}icture (SCAP) explanations \cite{Hinns2025}. SCAP explanations identify semantically meaningful segments within an image that, when altered (in our case, blurred), result in a change in the predicted class.
To do this, the image is split into segments by a pre-trained segmentation model, each of which is assigned a descriptive text label. Every segment is then blurred and the altered image is reclassified. If the altered image is classified differently, the altered segment is identified as a counterfactual.
Specifically, we use the DETR panoptic segmentation model \cite{carion2020}, fine-tuned on Microsoft’s COCO dataset \cite{lin2014}, which supports 133 possible labels. 
Integrating these labeled segments into a prompt is straightforward: “The image was initially classified as [Initial Class], however, when [Counterfactual Segment] was blurred, the model changed its classification to [Counterfactual Class]”.

Additionally, we use the SceneXplain plugin for GPT-4 \cite{SceneXplain}, which represents an image as a detailed text description. 
By adding this description to the prompt, we are able to provide a LLM with information about the image without providing the image itself.
Multi-modal LLMs such as GPT-4o could be greatly beneficial here, by removing the need for SceneXplain entirely. This would give more information to the LLM than a description of an image when generating narratives.
In our experiments, we use a pre-trained Vision Transformer (ViT) \cite{Dosovitskiy2020}, fine-tuned on Imagenet-1k, as the classifier we aim to explain. We then present narratives for images misclassified on the COCO dataset. 

We focus on misclassified images only as the counterfactuals for the correctly classified ones are trivial for these images. For example an image of a dog being correctly predicted to be a dog, has a trivial counterfactual segment to be the dog in the picture. 

\noindent \textbf{Tabular classification:}
For tabular classification, we focus on SHAP explanations, due to the popularity of SHAP for such data. This also complements the counterfactual based studies with the image data.

\noindent The prompt to generate \textit{SHAPstories} consists of several components. First, it introduces the dataset, outlining its target variable and its features. Such a data description, detailing the meaning of each feature, is commonly available. Second, it explains the functioning of a SHAP explanation and sets the context for the prediction by specifying the predicted class and whether the model's classification was accurate. Third, the LLM is asked to come up with a plausible, fluent story as to why the model could have predicted the specific outcome. Here, the LLM is explicitly required to use the highest absolute values from the SHAP explanation, explain the most important features only, end with a summary and limit its response to eight sentences maximum. Prioritising only the highest absolute values from the SHAP explanation addresses the limitation of SHAP, which assigns importance to all features \cite{Molnar2022}. Ending with a summary and limiting the response length can increase the persuasiveness of narratives, as discussed by Dahlstrom \cite{Dahlstrom2014}. Finally, the prompt provides a table which has the value and SHAP value of each feature of the data instance.  

\noindent For all tabular experiments, we train a Random Forest classifier using cross-validation and RandomSearchCV using the Python scikit-learn package \citep{scikit_learn2011}. This also allows the use of the Python implementation of treeSHAP \citep{Lundberg2017, Lundberg2020} for fast and accurate SHAP calculations.  

\begin{itemize}
    \item FIFA World Cup 2018 Man of The Match dataset, to predict whether a football team will have the `Man of the Match' winner in a FIFA 2018 world cup match, based on the team's statistics.
    \item Student Performance dataset, to predict whether a student in secondary education will pass or fail mathematics.
    \item German Credit Scoring Dataset, to predict creditworthiness of loan applicants based on the loan and applicant characteristics.
\end{itemize}
 These datasets are selected based on their features that allow semantic interpretation, their typical number of features (around 25 to 40), and because they originate from different domains (sports, education, financial intermediation). 

\noindent \textbf{Data and Code:} All experiments are run on Google Colab. The generation of GPT-4 responses for CFstories is performed in ChatGPT's playground, whilst SHAPstories use the API.
The trained models, prompts, data and a full implementation for SHAPstories used in our experiments can be found on the project \href{https://github.com/ADMAntwerp/XAIstories}{GitHub repository}.

\subsection{Survey design} \label{methodologySurvey}
\begin{figure}[tbh]
\includegraphics[width=14cm]{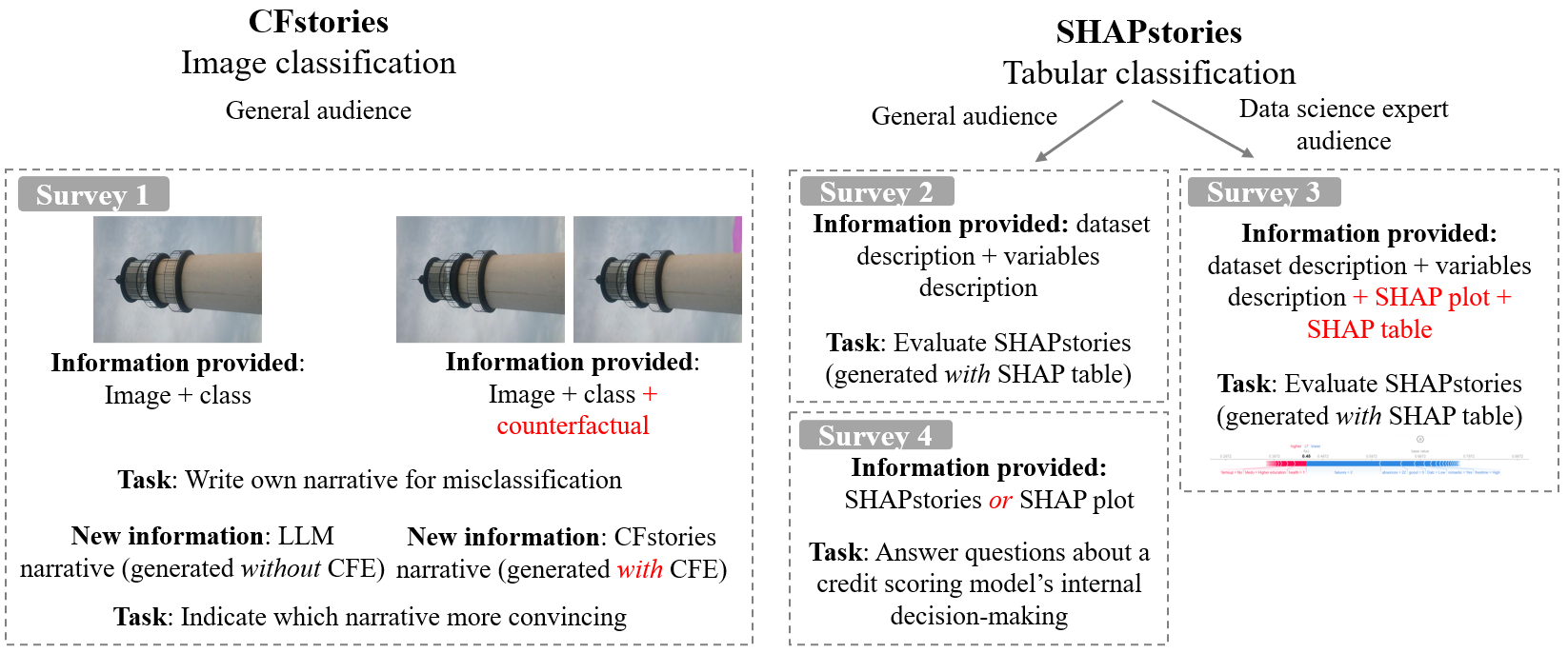}
\caption{Description of survey structure for CFstories (left) and SHAPstories (right)}
\label{fig:survey}
\end{figure}

We conducted four surveys, differing in the target audience (lay users versus data science experts), XAIstory (SHAPstory vs CFstory), and evaluation (user experience vs decision making and understanding). An overview of the structure of our surveys can be found in Figure~\ref{fig:survey}.

\noindent \textbf{Survey 1: User experience on CFstories by lay users.} The first survey aims at the evaluation of CFstories by a broader audience not trained in data science. We used Amazon Mechanical Turk (AMT). Respondents are provided images misclassified by an AI model. They are asked to come up with a two-line narrative, explaining what may have caused the misclassification. Following this, we provide the respondents with an alternative narrative, generated by CFstories, and ask them  to compare it with their own. 

\noindent \noindent \textbf{Survey 2: User experience on SHAPstories by lay users.} The second survey is set up to evaluate SHAPstories qualitatively. This survey again targets a broad audience, via AMT, and involves a two-round questionnaire. In each round, participants are presented an application for which an AI model made a correct or incorrect prediction. The respondents are provided with an overview of the dataset's features and the narrative generated by SHAPstories. Subsequently, participants are asked to assess whether the provided story offers a convincing interpretation of the model's decision.

\noindent  \textbf{Survey 3: User experience on SHAPstories by data scientists.} The third study also aims to assess SHAPstories qualitatively, but targets an audience with expertise in data science. This study is sent to data scientists recruited from our own network (excluding our own research group), without any reward for completion. The structure of the questions is the exact same as in the second survey, except that the data science experts are also given the SHAP force plot and SHAP values. Participants are asked whether they find the generated narratives convincing interpretations of the model's decisions and whether they consider the narratives to be a valuable addition to the SHAP force plot.

\noindent \textbf{Survey 4: User comprehension improvement with SHAPstories by lay users.} In the fourth survey we quantitatively evaluate whether SHAPstories can improve the understanding of AI decisions. The study is based on a credit scoring decision making setting where ground truth for the comprehension questions asked is available. In this survey, an introductory text explains that the respondent is asked to take the role of a banking employee and evaluate an AI model’s decision about the creditworthiness of a loan applicant. Respondents are presented a list of questions that assess their understanding of the AI model's internal functioning. The questions are shown in Table~\ref{table:methodology_Quantitative}, covering basic understanding as well as a hypothetical decision (question 4). Each of the questions is presented twice, once showing a SHAPstory and once showing a SHAP plot. Evaluation criteria are based on Huysmans et al. (2011) ~\cite{ huysmans11} who also assesses decision quality in an AI context. Since the ground truth is known for the first 3 questions, this setup allows us to evaluate the respondents' accuracy when answering the questions. We also record the time needed for the respondents to answer each question. A total of 8 questions are presented to the respondents (4 SHAP plots and 4 SHAPstories), where each question is about a different classification instance.

\begin{table}[H]
\begin{small}
\caption {SHAPstories' survey questions, describing which questions were asked to the respondents, both for SHAPstories and SHAP plots.}
\label{table:methodology_Quantitative}
\begin{singlespace}
\begin{tabular}{lp{12cm}}
\toprule
\textbf{Number} & \textbf{Question}                                                                                                                                                                                           \\ \midrule
1               & Which of the following one-sentence summaries best describes why this individual gets a \textbf{poor / good} creditworthiness assessment?                                                                            \\
2               & Imagine feature \textbf{X} and feature \textbf{Y} \textbf{increase / decrease}. How is this likely to affect the model's decision?                                                                                                     \\
3               & Select the statement that corresponds to the \textbf{SHAP story / plot}.                                                                                                                                        \\
4               & Position yourself in the position of a banking employee. Your bank is only accepting clients with above-average, high, creditworthiness. Which credit decision would you make based on the above \textbf{SHAP story / plot}? \\ \bottomrule
\end{tabular}
\end{singlespace}
\end{small}
\end{table}
\vspace{-2em}

\noindent Finally, in all SHAPstories surveys  (surveys 2-4), the respondents are asked to assess the narratives provided by SHAPstories along four dimensions. Following previous studies on the evaluation of human understanding of and trust in AI models \citep{Slack2023}, participants evaluated the following statements on a 1-5 Likert scale: 

\begin{itemize}
\itemsep-.70em 
    \item Ease: I felt the narratives were easy to interpret.
    \item Confidence: I felt more confident in my understanding of the prediction with the narratives than without the narratives.
    \item Speed: I felt I was able to arrive more rapidly at an understanding of the prediction with the narratives than without the narratives.
    \item Likeliness to use: Based on my experience with the narratives, I consider them a valuable addition to AI models, that I would like to use.  
\end{itemize}

\subsection{Survey respondents}

\noindent \textbf{CFstories:} For the user study on CFstories, the survey was sent to 160 respondents. After selecting only those respondents who filled in the complete questionnaire (122), passed the attention check (100), and after removal of the respondents who did not understand the task they were provided\footnote{We removed respondents for not properly engaging with the task—specifically, those who offered definitions instead of narratives, explained AI mechanisms, dismissed the model's output without explanation, or submitted random words for the reward.}, we ended up with a final sample of 82 surveys, in which 492 valid responses were recorded. 

\noindent \textbf{SHAPstories:} As noted above, we did three surveys with SHAPstories. The first two aimed at a qualitative assessment of SHAPstories. The non-data science audience that responded to this survey consisted of 70 respondents. After selecting only those respondents who passed the attention check and filled in the complete questionnaire, we ended up with a final sample of 66 surveys. For the data science expert audience, the survey on SHAPstories received full responses from 49 participants, with 36 of them successfully passing the attention check. The third survey examined SHAPstories from a more quantitative perspective, and was sent to 75 individuals. Of these individuals, 68 completed the full survey and passed the attention check. 

\begin{small}
\begin{table}[H]
\begin{singlespace}
\begin{longtable}{p{14cm}}
\caption {Example of a SHAPStory for the Student Performance Dataset.} \label{table:appendix_SHAPstories_question_student} \\
\hline
A certain student has the following features
\vspace{0.5em}

\begin{scriptsize}
\begin{center}
\begin{tabular}{lllll}
\hline
\textbf{Feature}     & \textbf{Value} &  & \textbf{Feature} & \textbf{Value} \\ \hline
Health                                   & 1                                  &        & Father's job\_teacher                & no                                 \\
Mother's education                       & higher educ                   &       & Reason  choose this school\_home   & no                                 \\
Family educational support               & no                                 &        & Father's job\_other                  & yes                                \\
Home to school travel time               & \textless 15 min.                  &        & Quality of family relationships      & average                            \\
Extra paid courses in math               & yes                                &                            & Mother's job\_services               & no                                 \\
Reason choose this school\_reputation & yes                                &                            & Mother's job\_health                 & no                                 \\
... & & & \\
                              \\ \hline                            
\end{tabular}
\end{center}
\end{scriptsize}

\vspace{1em}

The AI model predicted that the student will fail mathematics. This prediction is correct. The student failed mathematics. The following narrative provides an interpretation as to why the model could have made this prediction:

\vspace{0.5em}
\begin{footnotesize}
    
\textit{"The student in question has been going through a challenging time. Their health status is quite poor, which could be affecting their ability to focus on academics. Despite their mother's higher educational background, they do not receive educational support from their family. The student's proximity to the school, taking less than 15 minutes to travel, and the fact that they take extra paid classes hint at an effort to improve academically. [...]}
\vspace{0.5em}

\textit{In summary, while there are efforts visible from the student's side to excel academically, past failures, frequent absences, and an active social life might have influenced the AI model's prediction towards a higher probability of them failing in mathematics."}

\end{footnotesize}

\textbf{Do you think this story provides a convincing interpretation of the model's prediction?} [Answer options: Yes / No (please explain)]

\\ \hline 
\end{longtable}
\end{singlespace}
\end{table}
\end{small}

\section{Results} \label{Results}
We first provide some examples to illustrate different automatically generated narrative explanations, and then discuss the survey results. Note that we did not cherry pick results generated by LLMs, but rather took the first response we obtained. By repeating the prompt several times, and choosing among the answers, the results will likely improve further.

\subsection{XAIstories examples} \label{ResultsExamples}
\begin{footnotesize}
\begin{singlespace}
\begin{longtable}{p{2cm}p{2.25cm}p{2.25cm}p{3.5cm}p{3.5cm}} 
\caption {Overview of images and obtained narratives generated by LLMstories, which do not have access to the CF explanations (indicated in purple), and CFstories, which do.} \label{table:results_CFstories_images} \\
\hline
& \textbf{Image without CF}     &  \textbf{Image with CF}  & \textbf{LLMstories}  & \textbf{CFstories}   \\ \hline \endhead
Original classification: missile \newline CF: cloud 

& 
\vspace{0.25em}
\includegraphics[width=2.5cm, height=2.5cm]{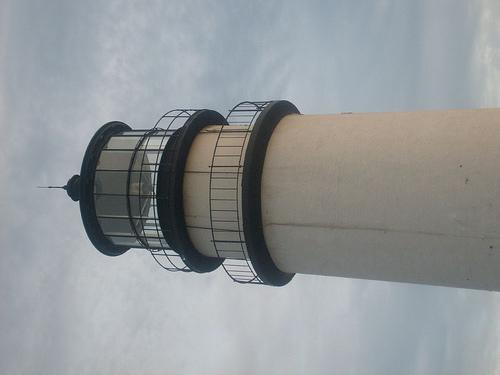}

 & 
\vspace{0.25em}
\includegraphics[width=2.5cm, height=2.5cm]{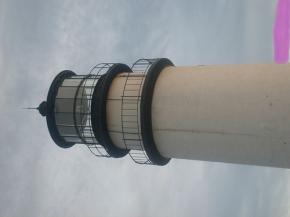}

&
\textit{The classifier might have mistaken the elongated shape and structure of the lighthouse against the sky as the silhouette of a missile.
}
&  \textit{The cloud's shape and position in relation to the lighthouse might resemble the trail of a missile launch, causing the classifier to misidentify the image.
}

\\




Original classification: unicycle \newline CF: person & 
\vspace{0.25em} \includegraphics[width=2.5cm, height=2.5cm]{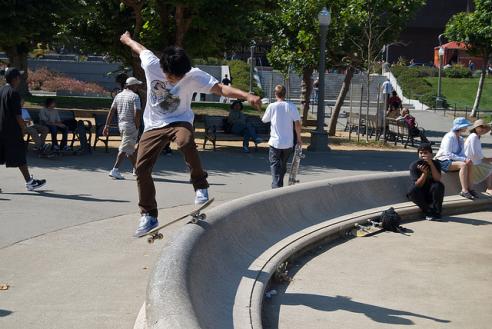}

& 
\vspace{0.25em} \includegraphics[width=2.5cm, height=2.5cm]{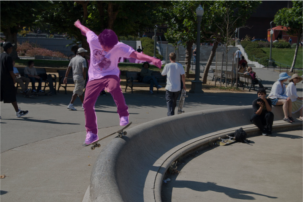}

&

\textit{The AI image classifier might have misclassified the image as a unicycle due to the skateboard's single wheel visibility and the skateboarder's body position, which could resemble a unicycle rider.}

& 
 \textit{The presence of a person, especially in dynamic poses, might lead the classifier to associate them with objects like unicycles due to similarity in balance and motion patterns often seen in unicycle performances.
}

 \\
\hline
\end{longtable}
\end{singlespace}
\end{footnotesize}

\noindent \textbf{CFstories:} Table~\ref{table:results_CFstories_images} shows two example images used to evaluate CFstories, along with the corresponding narratives generated by the CFstories. Presented with the counterfactuals, CFstories provide arguably better explanations with better reasoning. Baseline LLMstories, where the LLM is not provided the counterfactual explanations, are often unsuccessful in explaining the misclassifications as well.

\pagebreak

\begin{small}
\begin{table}[h]
\begin{singlespace}
\begin{longtable}{p{14cm}}
\caption {SHAPstory narrative generated for the SHAP plot found in Figure~\ref{fig:introexample}.}  \label{table:appendix_SHAPstories_question_loan} \\
\hline
A certain loan applicant has the following features:
\vspace{0.5em}

\begin{scriptsize}
\begin{center}
\begin{tabularx}{\textwidth}{lXlX}
\hline
\textbf{Feature} & \textbf{Value} & \textbf{Feature} & \textbf{Value} \\ \hline
Account Status & $<$ 0 DM & Property & real estate \\
Duration & 18 months & Age & 27 years \\
Duration in Current Address & $<$ 1 year & Other Loans & bank \\
Purpose & education & Housing & rent \\
Credit Amount & 750 DM & Existing Credits & 1 \\
Savings & $<$ 100 DM & Occupation & unemployed \\
Employment History & unemployed & Dependants & 3 or more dependants \\
Installment Rate & 4\% & Telephone & none \\
Sex and Marital Status & male : married & Foreign Worker & yes \\
Other Debtors & none & Credit History & existing credits paid back duly till now \\ \hline
\end{tabularx}
\end{center}
\end{scriptsize}

\vspace{1em}

The AI model predicted that the applicant should be denied a loan. The following narrative provides an interpretation as to why the model could have made this prediction:

\vspace{0.5em}
\begin{footnotesize}


\textit{Our individual is a young, unemployed person with a significant savings amount less than 100 DM (Deutsche Mark). They reside in their own house and have lived at their current address for less than a year. Their current account status indicates a balance less than 0 DM which appears as the largest negative contributor to the prediction. Furthermore, they recently took a loan amounting to 750 DM, with the purpose of education, from a bank where they have a credit history of existing credits paid off duly till now. However, the negative SHAP value for credit history suggests that the AI notes some risk associated with this factor. Employment history also weighs in heavily against them, as they are unemployed.} [...]
\vspace{0.5em}

\textit{In summary, while the individual did have some positive factors in their favor, their account status, employment status, savings amount, and recent credit amount tipped the balance towards more risk for the lender, causing the model to predict them as more likely 'not creditworthy'. The decision seems to underscore the AI's emphasis on an individual's financial stability and income sources when predicting creditworthiness.}

\end{footnotesize}

\\ \hline 
\end{longtable}
\end{singlespace}
\end{table}
\end{small}

\noindent \textbf{SHAPstories:} A SHAPstory example is shown in Table 4, based on the 
SHAP values example noted in Figure~\ref{fig:introexample} in the Introduction. The example is for a loan applicant who is predicted to have bad creditworthiness. The SHAPstories seem understandable, natural and sufficiently detailed, likely to be satisfactory for non-data science experts \citep{Chromik2021, Hoffman2023}.

\subsection{User surveys results}

\subsubsection{CFstories: Main results}
We compare CFstories and LLMstories along the dimensions of convincingness and speed, and relative to a baseline in which end users construct narratives themselves. We also measure how often CFstories and users' narratives include the counterfactuals. 

\noindent \textbf{Convincingness:} 
75.4\% of all respondents find CFstories as convincing, or more convincing, than their own narrative. For the baseline LLMstories, this number drops to 66.8\%. Both are statistically significant (relative to 50\%) at 1\% level for a one-tailed test on population proportion.

\noindent \textbf{Speed:} We also measure the time it takes for respondents to provide their responses. This allows us to compare the time to think of, and write a narrative, with the time they needed to simply read and evaluate the automatically generated narratives. On average, across all valid responses within the sample, respondents need 68 seconds to compose a narrative. In contrast, the average response time for evaluating the generated narratives is only 5 seconds, showing a significant speed advantage.

\noindent We also measure how often CFstories and users' narratives include the counterfactual explanations. While one would expect CFstories would always do so, there is no reason to guarantee the LLM would achieve this.  
In our data, CFstories narratives always included the given counterfactual, while  
only 79.7\% of the narratives provided by the respondents contained the counterfactual (or a synonym). 

In Table~\ref{table:results_CFstories_survey3}, we further examine the CFstories survey results by analysing each survey question separately. For all images, except image 3, more than half of the respondents find CFstories' narratives at least as convincing or even more convincing than their own narratives. It is important to note that variations between images exist. For question two, individuals exhibit the highest level of trust in CFstories' narrative. Here, the CFstories provided the following alternate story: \textit{``The presence of a person, especially in dynamic poses, might lead the classifier to associate them with objects like unicycles due to the similarity in balance and motion patterns often seen in unicycle performances"}, incorporating keywords that also emerged in respondents' explanations, such as balance, position, and the presence of a person.

\pagebreak

\noindent Finally, Table~\ref{table:results_CFstories_survey3} gives insights into why the respondents may favour their own narrative, when that happens. In the cases where the CF is already provided, individuals who prefer their own story often tend to include additional characteristics that could have led to the misclassification in their narrative. For instance, in the `missile' example, respondents added the orientation of the lighthouse as an additional reason for misclassification in their storytelling. One respondent described this as follows: \textit{``The image was misclassified because of the horizontal orientation of the lighthouse making it appear it is heading through clouds like a missile would"}. These kinds of explanations also arise for the other images. 

\subsubsection{SHAPstories: Main results}

\textbf{Qualitative results}\\
We studied subjective assessments of narratives along four dimensions: ease, confidence, speed, and likeliness to use. The findings are summarised in Table~\ref{table:results_SHAPstories_survey}.

\noindent \textit{Convincingness:} The general audience (survey 2) finds the narratives generated by SHAPstories more convincing interpretations of the model's predictions in 93.2\% of the cases. Non-data science experts, however, criticize the provided narratives for their limited set of attributes. From their viewpoint, this constraint leads to an incomplete portrayal of the factors influencing the AI model's predictions. This is due to the construction of the SHAPstories' prompt (focus on the top attributes only), and aligns with its goal of representing the most important information only. 

\noindent For the data science audience (survey 3), we find similar positive results. In $77.8\%$  of these evaluations, the data science experts agreed that SHAPstories convincingly explain the model's prediction. Furthermore, in $81.4\%$ of the cases, these experts agree that the narratives produced by SHAPstories serve as a valuable addition to the SHAP plots in explaining the AI model's predictions. 

\begin{landscape}
\thispagestyle{empty}
    
\begin{table}
\begin{threeparttable}
\footnotesize
\caption {CFstories' survey results, describing which narrative users found more convincing, and a summary of keywords that users included in their narratives.}
\label{table:results_CFstories_survey3}
\begin{tabular}{p{2.8cm}p{2.8cm}p{2.8cm}p{2.8cm}p{2.8cm}p{2.8cm}p{2.8cm}}
\hline \hline
 & \textbf{Question 1} 
 & \textbf{Question 2}  & \textbf{Question 3}  & \textbf{Question 4}  & \textbf{Question 5} & \textbf{Question 6} \\
 \hline
\vspace{0.5em} \textbf{Image} & \vspace{0.5em} \includegraphics[width=2.5cm, height=2.5cm]{figures/Lighthouse.jpg} \newline Class: missile \newline CF: cloud
 & \vspace{0.5em} \includegraphics[width=2.5cm, height=2.5cm]{figures/Skateboard.jpg} \newline Class: unicycle \newline CF: person & \vspace{0.5em} \includegraphics[width=2.5cm, height=2.5cm]{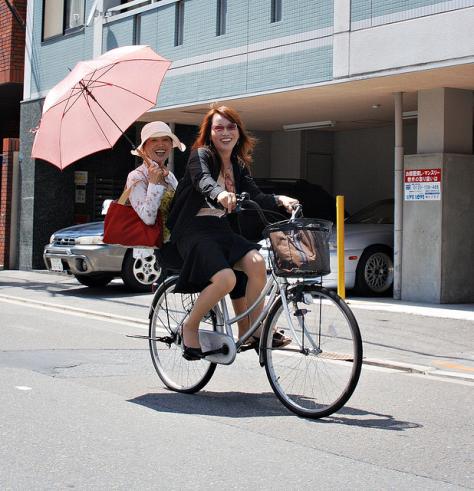} \newline Class: rickshaw \newline CF: person & \vspace{0.5em} \includegraphics[width=2.5cm, height=2.5cm]{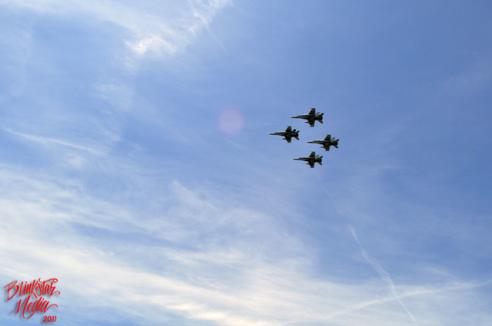}  \newline Class: goose \newline CF: sky & \vspace{0.5em} \includegraphics[width=2.5cm, height=2.5cm]{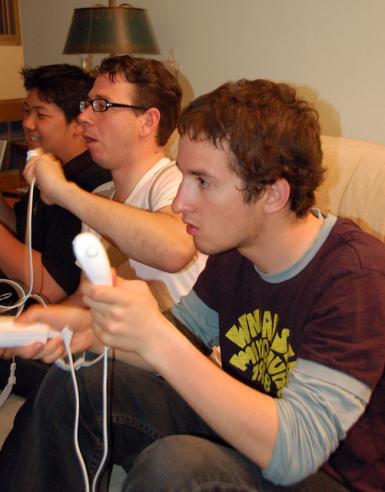} \newline Class: blow dryer \newline CF: person & \vspace{0.5em} \includegraphics[width=2.5cm, height=2.5cm]{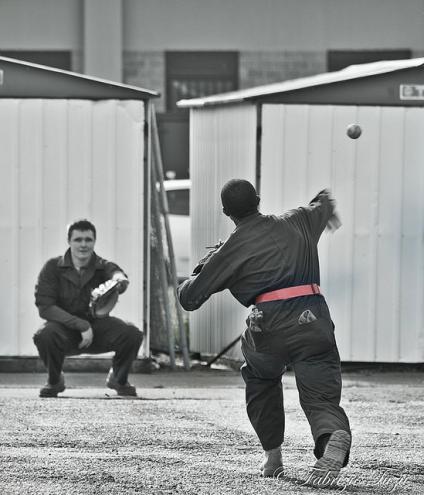} \newline Class: soccer ball \newline CF: person \\

\vspace{0.5em} \textbf{Results} & 

    \vspace{0.5em}
    \begin{tabular}{l|l}
    \hline \hline
    
    & (\%) \\
    Own & 33.3 \\
    Similar & 33.3 \\ 
    LLM & 33.3 \\
    \hline \hline

    \end{tabular}

     $p$-value: 0.0228

&

    \vspace{0.5em}
    \begin{tabular}{l|l}
    \hline \hline
    
    & (\%) \\
    Own & 9.8 \\
    Similar & 29.3 \\ 
    LLM & 61.0 \\
    \hline \hline

    \end{tabular}
    
    $p$-value: 0.0000

&

    \vspace{0.5em}
    \begin{tabular}{l|l}
    \hline \hline
    
    & (\%) \\
    Own & 42.9 \\
    Similar & 26.2 \\ 
    LLM & 31.0 \\
    \hline \hline

    \end{tabular}
    
    $p$-value: 0.1773

&

    \vspace{0.5em}
    \begin{tabular}{l|l}
    \hline \hline
    
    & (\%) \\
    Own & 32.4 \\
    Similar & 29.7 \\ 
    LLM & 37.8 \\
    \hline \hline

    \end{tabular}
    
    $p$-value: 0.0163

&

    \vspace{0.5em}
    \begin{tabular}{l|l}
    \hline \hline
    
    & (\%) \\
    Own & 17.5 \\
    Similar & 57.5 \\ 
    LLM & 25.0 \\
    \hline \hline

    \end{tabular}
    
    $p$-value: 0.0000

& 

    \vspace{0.5em}
    \begin{tabular}{l|l}
    \hline \hline
    
    & (\%) \\
    Own & 12.5 \\
    Similar & 42.5 \\ 
    LLM & 45.0 \\
    \hline \hline

    \end{tabular}
    
    $p$-value: 0.0000

\\

\vspace{0.25em} \textbf{Keywords: CF provided} &  

\footnotesize
\vspace{0.25em} clouds, (horizontal) orientation, shape, smoke/ gasses 

&
\footnotesize
\vspace{0.25em} wheels, person (position, balance) 

&
\footnotesize

\vspace{0.25em} wheeled vehicle, two people (passenger, second rider, behind driver), umbrella, shadow

&
\footnotesize

\vspace{0.25em} formation flying (symmetrical, leader), sky (blue, water resemblance, distant)  

&
\footnotesize

\vspace{0.25em} person (proximity to hair/face), cord/wire 

& 
\footnotesize

\vspace{0.25em} person (goalkeeper, kneeling, dribbling)  

\\

\vspace{0.25em} \textbf{Keywords: CF not provided}&  

\footnotesize
\vspace{0.25em} (horizontal/ sideways) orientation, shape (tall, thin, pointed, cylindrical), color, antenna

&
\footnotesize
\vspace{0.25em} person (posture, balance, position), surface (curve), background (circle)

&
\footnotesize

\vspace{0.25em} umbrella, rickshaw (wheels), two people (passenger, driver)

&
\footnotesize

\vspace{0.25em} formation flying (flock), plane (size, color)

&
\footnotesize

\vspace{0.25em} object (shape, cord/wire, handheld), person (proximity to hair/face)

& 
\footnotesize

\vspace{0.25em} person (stance/position, goalkeeper, kick), ball (flying), background (field, grass)

\\ \hline \hline
\end{tabular}
\begin{tablenotes}
\footnotesize
\item Note: The  displayed $p$-value are results for the tests on population proportion, grouping together individuals who indicated CFstories' narrative to be of similar quality or better than their own narrative. 
\end{tablenotes}
\end{threeparttable}
\end{table}

\end{landscape}

\begin{center}
\begin{singlespace}
\begin{threeparttable}
\caption {SHAPstories: results for general audience (left), data scientist own usage (middle) and data scientist usage for general audiences (right).}
\label{table:results_SHAPstories_survey}
\begin{small}
\begin{tabular}{l|ccc}
\hline
                           &                                                                                                                                                                                                                                                                                                                              \begin{tabular}[c]{@{}c@{}}SHAPstories: \\ general \\ audience\end{tabular} & \begin{tabular}[c]{@{}c@{}}SHAPstories: \\ data science \\ audience own \\ usage\end{tabular} & \begin{tabular}[c]{@{}c@{}}SHAPstories: \\ data science \\ audience: general \\ audience usage\end{tabular} \\ \hline
\textbf{Convincingness}    & \cellcolor[HTML]{77C68C}93.2\%  $^{***}$                                         & \multicolumn{2}{c}{\cellcolor[HTML]{8ED0A0}77.8\% $^{***}$}                                                                                                                                                  \\
\textbf{Ease}                                                                                                                                                                                                                                                                                                                                                                                                                                                                       & \cellcolor[HTML]{79C78E}92.4\% $^{***}$                                             & \cellcolor[HTML]{DBEFE2}61.1\% $^*$                                                            & \cellcolor[HTML]{63BE7B}91.7\% $^{***}$                                                                            \\
\textbf{Confidence}                                                                                                                                                                                                                                                                                                                                                                                                                                                                       & \cellcolor[HTML]{9ED6AE}79.7\% $^{***}$                                           & \cellcolor[HTML]{FCFCFF}38.9\%                                                            & \cellcolor[HTML]{63BE7B}91.7\% $^{***}$                                                                       \\
\textbf{Speed}                                                                                                                                                                                                                                                                                                                                                                                                                   & \cellcolor[HTML]{88CD9B}87.5\%  $^{***}$                                         & \cellcolor[HTML]{FCFCFF}44.4\%                                                            & \cellcolor[HTML]{96D3A7}88.9\% $^{***}$                                                                        \\
\textbf{Likeliness to use}                                                                                                                                     & \cellcolor[HTML]{7AC88F}92.2 \% $^{***}$                                           & \cellcolor[HTML]{B9E1C5}61.1\% $^*$                                                            & \cellcolor[HTML]{96D3A7}83.3\% $^{***}$                                                                         \\ \hline
\end{tabular}
\end{small}
 \begin{tablenotes}
     \begin{small}
     \item \textit{Note}: For all results we test if the obtained result is statistically significant: *** $p < 0.01$, ** $p < 0.05$, * $p < 0.1$
     \end{small}
 \end{tablenotes}
\end{threeparttable}
\end{singlespace}
\end{center}

\vspace{0.5cm}

\noindent \textit{Ease, confidence, speed and likeliness to use:} Table~\ref{table:results_SHAPstories_survey} examines four dimensions along which the respondents are asked to assess SHAPstories: 92.42\% of the non-data scientist respondents agree or strongly agree with the statement that SHAPstories are easy to interpret, and 92.2\% finds SHAPstories a valuable addition to AI models, that they are likely to use. Also, 87.5\% of the non-data scientist respondents indicate that SHAPstories increase the speed with which they understand the AI model's prediction. Overall, these results suggest that SHAPstories are well-received in terms of user-friendliness, confidence, and efficiency, which bodes well for adoption and continued usage. 

\noindent The second and third columns of Table~\ref{table:results_SHAPstories_survey} examine the same dimensions, but only for the data science experts. In this context, data science experts are asked to evaluate SHAPstories both for their own use, and for a broader non-expert audience. When considering its utility for their own usage, a majority (61.1\%) of data science experts agree on the ease of use that SHAPstories provide. The data science experts also evaluate SHAPstories very positively having in mind the usage for a general audience. These experts indicate in 91.7\% of the cases that SHAPstories will contribute to a non-specialist's ease and confidence in understanding an AI model's predictions. This is also reflected in the high likelihood that data scientists will use SHAPstories for a non-data scientists audience (83.3\%). 

In our fourth survey, we verified the consistency of our findings and additionally asked respondents to evaluate SHAP plots. The results corroborate earlier findings: SHAPstories were preferred in terms of ease, confidence, speed, and likeliness to use by respectively $79.0\%, 79.4\%, 72.1\%$ and $85.3\%$.

\noindent \textit{Qualitative feedback:}
Finally, we gathered qualitative open ended feedback from the data science experts. We provide representative quotes indicating the potential of SHAPstories.

\noindent  \textit{``They [SHAPstories] could be exceedingly helpful when it comes to evaluating model outputs by business users so that they can trust the model's output more easily. It could also help when formulating a rejection to a specific customer."} \newline 18 respondents emphasize the utility of SHAPstories in scenarios where explanations of individual predictions are required for non-data science audiences. Examples include popular science communication, business communication, and storytelling for medical experts. Furthermore, they also highlight its effectiveness when multiple narratives are necessary.

\noindent  \textit{``It is easier to evaluate (and/or correct/modify) the correctness of a story against the numbers and the plots than it is to go through the numbers and construct the story from scratch."} \newline This quote was phrased only by one respondent, and indicates that SHAPstories can address some limitations of SHAP, namely that SHAP attributes significance to all features \citep{Molnar2022} and does not have an established protocol for interpretation \citep{Kumar2020}. 

\noindent  \textit{``It [SHAPstories] goes beyond explanation"} or \textit{``[the narrative is] postulating beyond the facts."} \newline
Five respondents emphasize that SHAPstories makes underlying assumptions that are not necessarily correct, a limitation we discuss below. 

\noindent  \textit{``The added value is rather low as it mostly repeats the information from the plot."} \newline Six other respondents do not immediately see the added value of SHAPstories, stating that SHAP plot and tables already convey the same information. These respondents also mentioned that the SHAP plot and table provide more intuitive and efficient explanations. Earlier results, however, indicate that most of the data science experts refrain from such a critique when they have a non-data science audience in mind.

\textbf{Quantitative results}\\
Survey 4 studies to what extent the users' understanding of explanations are accurate. Accuracy is based on whether the respondents' answers to the questions in Table~\ref{table:methodology_Quantitative} agree with the correct answers to the questions -- which are objective, as they are based on the machine learning model's behavior, and unique for questions 1-3. Regarding decision question 4 in Table~\ref{table:methodology_Quantitative}, we measure whether the respondent's decision, while subjective, is the same as that of the machine learning model. A priori there is no reason why that would be the case. We also measure how long it takes for them to make a decision. The context is credit scoring. 

\noindent \textit{Accuracy:} 
Table \ref{table:results_survey_accuracy} shows the percentage of  correctly answered questions. Accuracy levels differ significantly depending on the question asked. For all first three questions, which ask respondents about their understanding of the internal-decision making of the model, SHAPstories scores the best. The differences are statistically significant (p-value $<$ 0.05) for the first two questions. The difference relative to SHAP plots is the largest (50\%) for question 1. SHAP plots, on the other hand, perform rather poorly indicating that the respondents do not understand well how the model has made a decision. The fourth question has the lowest score for both SHAPstories and SHAP plots, indicating that although people make different decisions than the model, they still understand the rationale of the model's decision as indicated by the first three questions. 

In line with all earlier results, these results further support that SHAPstories more accurately summarize and help users understand AI decisions.


\begin{table}[htb]
\begin{small}
\caption {SHAPstories' survey results, describing the proportion of respondents correctly answering questions. Results reported separately for SHAPstories and SHAP plots, with p-values from a one-sided test on the population proportion. Significant results are indicated in \textbf{bold}.}
\label{table:results_survey_accuracy}
\begin{tabular}{cccccc}
\hline
\textbf{} & \textbf{Obs} & \textbf{SHAPstories} & \textbf{SHAP plots} & \textbf{Difference} & \textbf{p-value} \\
\hline
Question 1        & 68           & \textbf{83.8\%}                    & 33.8\%                   & 50\%                & \textbf{0.000}            \\
Question 2        & 68           & \textbf{64.7\%}                    & 48.5\%                   & 16.2\%              & \textbf{0.029}            \\
Question 3        & 68           & 67.6\%                    & 61.7\%                   & 5.9\%               & 0.237            \\
\textit{Question 4}        & \textit{68}           & \textit{44.1\%}                    & \textit{44.1\%}                 & \textit{0\%}                 & \textit{0.500}           
\\ \hline
\end{tabular}
\end{small}
\end{table}

\noindent \textit{Speed:}
The time (measured in seconds) needed for respondents to answer questions are on average consistently around 1 minute, and do not significantly favour the stories or plots.
On average, respondents are 10.6\% slower when responding to questions based on SHAPstories. This increase in response time, however, improves the decision quality. To study this relationship further, we run two Poissons regressions, where accuracy and the confidence in their own answer serve as dependent variables. These are regressed on the time respondents need to answer a question. Our regression analysis shows a statistically significant positive relationship between the time taken to answer questions and accuracy for SHAPstories, suggesting that the additional time spent reading narratives significantly improves decision quality. Each second spent on narratives increased accuracy (coefficient = $0.00059, p < 0.05$) without significantly affecting respondents' confidence in their answers (coefficient = $-0.00014, p > 0.1$). In contrast, for SHAP plots, each additional second did not significantly increase accuracy (coefficient = $0.00061, p > 0.1$) but significantly reduced the respondents' confidence in their answers (coefficient = $-0.00045, 0.5 < p < 0.1$), indicating that more time on plots might confuse respondents rather than help them. These findings highlight the distinct impacts of narrative versus plot-based explanations on decision-making processes.

\section{Discussion \& Conclusion} \label{conclusion}

As machine learning models continue to grow in complexity, there is an increasing need for techniques that can explain model predictions to a diverse range of end users. However, current state-of-the-art methods in XAI lack the ability to provide a narrative to the users. In this paper, we presented XAIstories as a novel technique to help further bridge the gap between XAI and human understanding. 

\noindent Our experiments show striking results: in an image classification study, CFstories are indicated as providing at least equally convincing narratives as the respondents' own narratives in more than 75\% of the cases. Additionally, CFstories allows for an order of magnitude time gain in understanding AI models and has compelling accuracy in transforming a counterfactual to a natural language narrative. In tabular classification, over 90\% of non-data scientists and over 75\% of data scientists agreed that SHAPstories provides a convincing interpretation of an AI model's prediction. Our results also indicate that SHAPstories enables people to develop a better understanding of how AI decisions are made (study 4). Furthermore, our results suggest that SHAPstories is well-received in terms of confidence and speed gains by the general audience. 

\noindent Our results are subject to several limitations. First, our surveys are distributed only among a general audience and data science experts; we did not specifically target domain experts. 
An assessment of XAIstories through in-domain validation cases, with domain experts, is an interesting avenue for future research. Additionally, our experiments are exclusively centered on GPT-4 as a LLM for providing storytelling in the context of XAI methods. Given our findings and the anticipated improvement in the performance of LLMs in general, we expect our results to further improve in the future. Finally, as indicated by some surveyed data scientists, the narratives inherently involve certain assumptions, much like when a human crafts a story from explanations. Articulating and assessing these assumptions, and more broadly assessing XAIstories, represents another interesting direction for future research. 

\paragraph{Declaration of generative AI and AI-assisted technologies in the writing process}
During the preparation of this work the authors used GPT-4 to run the analyses, as detailed in the text, and chatGPT as a writing assistant to rephrase sentences. After using this tool, the authors reviewed and edited the content as needed and take full responsibility for the content of the publication.

\textbf{Acknowledgements} \\
This research received funding from the Flemish Government under the ``Onderzoeksprogramma Artificiële Intelligentie (AI) Vlaanderen''.
This research received funding from the AXA Joint Research Initiative (JRI).

\bibliographystyle{elsarticle-num} 
\bibliography{cas-refs}

@article{adadi2018,
  title={Peeking inside the black-box: a survey on explainable artificial intelligence (XAI)},
  author={Adadi, Amina and Berrada, Mohammed},
  journal={IEEE access},
  volume={6},
  pages={52138--52160},
  year={2018},
  publisher={IEEE}
}

@article{bullock2021,
  title={Narratives are persuasive because they are easier to understand: examining processing fluency as a mechanism of narrative persuasion},
  author={Bullock, Olivia M and Shulman, Hillary C and Huskey, Richard},
  journal={Frontiers in Communication},
  volume={6},
  pages={719615},
  year={2021},
  publisher={Frontiers Media SA}
}

@inproceedings{chromik2021,
  title={Human-XAI interaction: a review and design principles for explanation user interfaces},
  author={Chromik, Michael and Butz, Andreas},
  booktitle={IFIP Conference on Human-Computer Interaction},
  pages={619--640},
  year={2021},
  organization={Springer}
}

@article{dahlstrom2014,
  title={Using narratives and storytelling to communicate science with nonexpert audiences},
  author={Dahlstrom, Michael F},
  journal={Proceedings of the national academy of sciences},
  volume={111},
  number={supplement\_4},
  pages={13614--13620},
  year={2014},
  publisher={National Academy of Sciences}
}

@article{dosovitskiy2020,
  title={An image is worth 16x16 words: Transformers for image recognition at scale},
  author={Dosovitskiy, Alexey and Beyer, Lucas and Kolesnikov, Alexander and Weissenborn, Dirk and Zhai, Xiaohua and Unterthiner, Thomas and Dehghani, Mostafa and Minderer, Matthias and Heigold, Georg and Gelly, Sylvain and others},
  journal={arXiv preprint arXiv:2010.11929},
  year={2020}
}

@article{gohel2021,
  title={Explainable AI: current status and future directions},
  author={Gohel, Prashant and Singh, Priyanka and Mohanty, Manoranjan},
  journal={arXiv preprint arXiv:2107.07045},
  year={2021}
}

@article{hoffman2023,
  title={Measures for explainable AI: Explanation goodness, user satisfaction, mental models, curiosity, trust, and human-AI performance},
  author={Hoffman, Robert R and Mueller, Shane T and Klein, Gary and Litman, Jordan},
  journal={Frontiers in Computer Science},
  volume={5},
  pages={1096257},
  year={2023},
  publisher={Frontiers Media SA}
}

@inproceedings{kumar2020,
  title={Problems with Shapley-value-based explanations as feature importance measures},
  author={Kumar, I Elizabeth and Venkatasubramanian, Suresh and Scheidegger, Carlos and Friedler, Sorelle},
  booktitle={International conference on machine learning},
  pages={5491--5500},
  year={2020},
  organization={PMLR}
}

@article{lundberg2017,
  title={A unified approach to interpreting model predictions},
  author={Lundberg, Scott M and Lee, Su-In},
  journal={Advances in neural information processing systems},
  volume={30},
  year={2017}
}

@article{lundberg2020,
  title={From local explanations to global understanding with explainable AI for trees},
  author={Lundberg, Scott M and Erion, Gabriel and Chen, Hugh and DeGrave, Alex and Prutkin, Jordan M and Nair, Bala and Katz, Ronit and Himmelfarb, Jonathan and Bansal, Nisha and Lee, Su-In},
  journal={Nature machine intelligence},
  volume={2},
  number={1},
  pages={56--67},
  year={2020},
  publisher={Nature Publishing Group UK London}
}

@article{kim2020,
  title={Transparency and accountability in AI decision support: Explaining and visualizing convolutional neural networks for text information},
  author={Kim, Buomsoo and Park, Jinsoo and Suh, Jihae},
  journal={Decision Support Systems},
  volume={134},
  pages={113302},
  year={2020},
  publisher={Elsevier}
}

@article{Kayande09,
	author = {U. Kayande and A. {De Bruyn} and G. L. Lilien and A. Rangaswamy and G. H. {van Bruggen}},
	title = {How Incorporating Feedback Mechanisms in a {DSS} Affects DSS Evaluations},
journal = {Information Systems Research},
volume = {20},
year = {2009},
pages = {527--546},
issue = {4}
}

@article{Martens2014,
author = {Martens, David and Provost, Foster},
title = {Explaining Data-Driven Document Classifications},
year = {2014},
issue_date = {March 2014},
publisher = {Society for Information Management and The Management Information Systems Research Center},
address = {USA},
volume = {38},
number = {1},
journal = {MIS Quarterly},
month = {mar},
pages = {73–100},
numpages = {28}
}

@book{martens2022data,
  title={Data Science Ethics: Concepts, Techniques, and Cautionary Tales},
  author={Martens, David},
  isbn={9780192847270},
  year={2022},
  publisher={Oxford University Press}
}

@book{Molnar2022,
  title      = {Interpretable Machine Learning},
  author     = {Christoph Molnar},
  year       = {2022},
  subtitle   = {A Guide for Making Black Box Models Explainable},
  edition    = {2},
  url        = {https://christophm.github.io/interpretable-ml-book}
}

@inproceedings{lin2014,
  title={Microsoft coco: Common objects in context},
  author={Lin, Tsung-Yi and Maire, Michael and Belongie, Serge and Hays, James and Perona, Pietro and Ramanan, Deva and Doll{\'a}r, Piotr and Zitnick, C Lawrence},
  booktitle={European conference on computer vision},
  pages={740--755},
  year={2014},
  organization={Springer}
}

@article{yang2023,
  title={Survey on explainable AI: From approaches, limitations and applications aspects},
  author={Yang, Wenli and Wei, Yuchen and Wei, Hanyu and Chen, Yanyu and Huang, Guan and Li, Xiang and Li, Renjie and Yao, Naimeng and Wang, Xinyi and Gu, Xiaotong and others},
  journal={Human-Centric Intelligent Systems},
  volume={3},
  number={3},
  pages={161--188},
  year={2023},
  publisher={Springer}
}

@inproceedings{ribeiro2016,
  title={" Why should i trust you?" Explaining the predictions of any classifier},
  author={Ribeiro, Marco Tulio and Singh, Sameer and Guestrin, Carlos},
  booktitle={Proceedings of the 22nd ACM SIGKDD international conference on knowledge discovery and data mining},
  pages={1135--1144},
  year={2016}
}

@misc{SceneXplain,
    author= {Jina},
    year  = {2023},
    title = {SceneXplain},
    note  = {\url{https://scenex.jina.ai/}, 
             Last accessed on 2023-07-20},
}

@article{scikit_learn2011,
 title={Scikit-learn: Machine Learning in {P}ython},
 author={Pedregosa, F. and Varoquaux, G. and Gramfort, A. and Michel, V.
         and Thirion, B. and Grisel, O. and Blondel, M. and Prettenhofer, P.
         and Weiss, R. and Dubourg, V. and Vanderplas, J. and Passos, A. and
         Cournapeau, D. and Brucher, M. and Perrot, M. and Duchesnay, E.},
 journal={Journal of Machine Learning Research},
 volume={12},
 pages={2825--2830},
 year={2011}
}

@article{slack2023,
  title={Explaining machine learning models with interactive natural language conversations using TalkToModel},
  author={Slack, Dylan and Krishna, Satyapriya and Lakkaraju, Himabindu and Singh, Sameer},
  journal={Nature Machine Intelligence},
  volume={5},
  number={8},
  pages={873--883},
  year={2023},
  publisher={Nature Publishing Group UK London}
}

@article{vermeire2022,
  title={Explainable image classification with evidence counterfactual},
  author={Vermeire, Tom and Brughmans, Dieter and Goethals, Sofie and De Oliveira, Raphael Mazzine Barbossa and Martens, David},
  journal={Pattern Analysis and Applications},
  volume={25},
  number={2},
  pages={315--335},
  year={2022},
  publisher={Springer}
}

@article{wachter2017,
  title={Counterfactual explanations without opening the black box: Automated decisions and the GDPR},
  author={Wachter, Sandra and Mittelstadt, Brent and Russell, Chris},
  journal={Harv. JL \& Tech.},
  volume={31},
  pages={841},
  year={2017},
  publisher={HeinOnline}
}

@article{delange2022,
  title={Explainable AI for credit assessment in banks},
  author={De Lange, Petter Eilif and Melsom, Borger and Venner{\o}d, Christian Bakke and Westgaard, Sjur},
  journal={Journal of Risk and Financial Management},
  volume={15},
  number={12},
  pages={556},
  year={2022},
  publisher={MDPI}
}

@misc{DeOliveira2023,
Author = {Raphael Mazzine Barbosa de Oliveira and Sofie Goethals and Dieter Brughmans and David Martens},
Title = {Unveiling the Potential of Counterfactuals Explanations in Employability},
Year = {2023},
Eprint = {arXiv:2305.10069},
}

@incollection{graesser2003,
  title={How does the mind construct and represent stories?},
  author={Graesser, Arthur C and Olde, Brent and Klettke, Bianca},
  booktitle={Narrative impact},
  pages={229--262},
  year={2003},
  publisher={Psychology Press}
}

@article{miller2017,
  title={Explainable AI: Beware of inmates running the asylum or: How I learnt to stop worrying and love the social and behavioural sciences},
  author={Miller, Tim and Howe, Piers and Sonenberg, Liz},
  journal={arXiv preprint arXiv:1712.00547},
  year={2017}
}

@misc{Provost2020, 
title={Machine learning in action for Compass’s likely-to-sell recommendations}, 
journal={Medium}, 
publisher={Compass True North}, 
author={D N, Varun and Ipeirotis, Panos and Provost, Foster}, 
year={2020}, 
month={Nov}
}

@inproceedings{walke2023,
  title={Artificial intelligence explainability requirements of the AI act and metrics for measuring compliance},
  author={Walke, Fabian and Bennek, Lars and Winkler, Till J},
  booktitle={International Conference on Wirtschaftsinformatik},
  pages={113--129},
  year={2023},
  organization={Springer}
}

@article{wulf2024,
  title={“Please understand we cannot provide further information”: evaluating content and transparency of GDPR-mandated AI disclosures},
  author={Wulf, Alexander J and Seizov, Ognyan},
  journal={AI \& SOCIETY},
  volume={39},
  number={1},
  pages={235--256},
  year={2024},
  publisher={Springer}
}

@article{zhou2023,
  title={A user-centered explainable artificial intelligence approach for financial fraud detection},
  author={Zhou, Ying and Li, Haoran and Xiao, Zhi and Qiu, Jing},
  journal={Finance Research Letters},
  volume={58},
  pages={104309},
  year={2023},
  publisher={Elsevier}
}

@inproceedings{hinns2025,
  title={Exposing Shortcuts in Image Classification by Aggregating Counterfactuals},
  author={Hinns, James and Martens, David},
  booktitle={International Joint Conference on Computational Intelligence},
  pages={451--466},
  year={2025},
  organization={Springer}
}

@article{bubeck2023,
  title={Paper Review:'Sparks of Artificial General Intelligence: Early experiments with GPT-4'},
  author={Bubeck, S{\'e}bastien and Chandrasekaran, Varun and Eldan, Ronen and Gehrke, Johannes and Horvitz, Eric and Kamar, Ece and Lee, Peter and Lee, Yin Tat and Li, Yuanzhi and Lundberg, Scott and others},
  year={2023}
}

@article{yang2024,
  title={Harnessing the power of llms in practice: A survey on chatgpt and beyond},
  author={Yang, Jingfeng and Jin, Hongye and Tang, Ruixiang and Han, Xiaotian and Feng, Qizhang and Jiang, Haoming and Zhong, Shaochen and Yin, Bing and Hu, Xia},
  journal={ACM Transactions on Knowledge Discovery from Data},
  volume={18},
  number={6},
  pages={1--32},
  year={2024},
  publisher={ACM New York, NY}
}

@inproceedings{carion2020,
  title={End-to-end object detection with transformers},
  author={Carion, Nicolas and Massa, Francisco and Synnaeve, Gabriel and Usunier, Nicolas and Kirillov, Alexander and Zagoruyko, Sergey},
  booktitle={European conference on computer vision},
  pages={213--229},
  year={2020},
  organization={Springer}
}

@article{dssstorey2024,
  title={The design of human-artificial intelligence systems in decision sciences: A look back and directions forward},
  author={Storey, Veda C and Hevner, Alan R and Yoon, Victoria Y},
  journal={Decision Support Systems},
  volume={182},
  pages={114230},
  year={2024},
  publisher={Elsevier}
}

@article{huysmans11,
  title={An empirical evaluation of the comprehensibility of decision table, tree and rule based predictive models},
  author={Huysmans, Johan and Dejaeger, Karel and Mues, Christophe and Vanthienen, Jan and Baesens, Bart},
  journal={Decision Support Systems},
  volume={51},
  number={1},
  pages={141--154},
  year={2011},
  publisher={Elsevier}
}

\end{document}